\documentclass[11pt,a4paper]{article}
\usepackage{authblk}

\usepackage[hyphens]{url}

\usepackage[hyperref]{emnlp2020}
\usepackage{times}
\usepackage{latexsym}

\usepackage{subcaption}
\usepackage{graphicx}
\usepackage{booktabs}
\usepackage{multirow}
\usepackage{amsfonts}
\usepackage{balance}
\usepackage{titlesec}

\usepackage[hyphenbreaks]{breakurl}

\usepackage{microtype}

\aclfinalcopy 
\def\aclpaperid{1390} 

\usepackage{caption}
\title{Conversational Document Prediction to Assist Customer Care Agents}

\author{
Jatin Ganhotra,\textsuperscript{\rm 1}
Haggai Roitman,\textsuperscript{\rm 2\thanks{\hspace{0.2cm}Work done when author was at IBM Research} }
Doron Cohen,\textsuperscript{\rm 1}
Nathaniel Mills,\textsuperscript{\rm 1}
Chulaka Gunasekara,\textsuperscript{\rm 1}\\
Yosi Mass,\textsuperscript{\rm 1}
Sachindra Joshi,\textsuperscript{\rm 1}
Luis Lastras,\textsuperscript{\rm 1}
David Konopnicki\textsuperscript{\rm 1}\\
\\
\textsuperscript{\rm 1}IBM Research,
\textsuperscript{\rm 2}eBay Research}

\affil[]{\textit {\{jatinganhotra, wnm3, lastrasl\}@us.ibm.com, hroitman@ebay.com, chulaka.gunasekara@ibm.com},\\ \textit{\{doronc, yosimass, davidko\}@il.ibm.com,  jsachind@in.ibm.com}}

\date{}

\begin{document}

\maketitle
\begin{abstract}

A frequent pattern in customer care conversations is the agents responding with appropriate webpage URLs that address users' needs. We study the task of predicting the documents that customer care agents can use to facilitate users' needs. We also introduce a new public dataset\footnote{The Twitter dataset is available at:\\ \url{https://github.com/IBM/twitter-customer-care-document-prediction}} which supports the aforementioned problem. Using this dataset and two others,  we investigate state-of-the-art deep learning (DL) and information retrieval (IR) models for the task. We also analyze the practicality of such systems in terms of inference time complexity. Our results show that an hybrid IR+DL approach provides the best of both worlds. 

\end{abstract}

\section{Introduction}
\label{sec:introduction}

Customer care (CC) agents play a crucial role as an organization's main representatives to the public. Our work is motivated by the observation that, in many conversations between CC agents and users, the former tend to provide links to documents that may help resolve user issues. This is a prevalent pattern that is found in around 5-9\% of all customer care conversations in multiple domains that we have reviewed.
To identify such documents, the agents manually extract the keywords from the conversation and search over their customer service knowledge base \cite{habibi2015keyword, ferreira2019classy}. Table \ref{tab:dataset-samples} shows a 
conversation where the agent provides a URL\footnote{The terms URL and document are used interchangeably.} to the user.

\begin{table}[hbt!]
\centering
\footnotesize
\begin{tabular}{p{0.2cm}p{6.5cm}}
\toprule
\textbf{Dialogue}
&             \\
\midrule
\textbf{U}: & My virtual keyboard seems to float in the screen. Not sure how to undo what I just did. Can you help me please? \\
\textbf{A}: & We’re happy to help. To start, let us know which device you’re working with, and the OS version installed on it. \\
\textbf{U}: & It is an iPad \\
\textbf{A}: & Ok, to check version, tap Settings; General ; About. \\
\textbf{U}: & It's iPad 4, 11 inch - model A1934 \\
\textbf{A}: & Thank you. This article can help with how to merge a split keyboard and move the keyboard for an iPad: \url{https://support.apple.com/en-us/HT207521}. Let me know if this helps \\

\bottomrule
\end{tabular}
\caption{Sample dialog from Twitter where the Agent (\textbf{A}) utterance includes a URL to the User (\textbf{U}) query.}
\label{tab:dataset-samples}
\end{table}

\begin{table*}[ht]
\centering
\footnotesize
\begin{tabular}{lccc}
\toprule
\textbf{Metric} & \textbf{Twitter} & \textbf{Mac-Support} & \textbf{Telco-Support} \\
\midrule

\# of dialogs        & 13500 & 83436 & 1634562 \\
\# of dialogs with an URL in agent utterance & 13500 & 10470 & 99733\\
\# of dialogs with an in-domain URL in agent utterance  & 13500 & 7531 & 84126 \\
\# of dialogs with an in-domain valid URL in agent utterance & 11025 & 4611 & 48565 \\
Valid/Total \# of unique URLs   &  2004 / 3585 & 522 / 1130 & 318 / 1203 \\
Avg./Max. \# of turns per dialogue  &  1.2 / 8.0  & 8.2 / 80.5 & 9.9 / 75.0 \\ 
Avg./Max. dialog length (in tokens)  & 40.5 / 503 & 145.8 / 1481 & 334.5 / 5390 \\
Avg./Max. URL content length (in tokens)  &  537.4 / 20492  & 311.6 / 7765 & 877.2 / 7889 \\
vocabulary size                     &  11646 & 10454 & 34099 \\
train/dev/test split size  &  10000 / 525 / 500  & 3677 / 467 / 467 & 38850 / 4857 / 4858 \\
\bottomrule
\end{tabular}
\caption{The overall statistics of the three datasets.}
\label{tab:data:overall}

\vspace{-5mm}
\end{table*}

Although responding with URLs is a common pattern, automating this process to aid the agents remains underexplored in the literature. This task of \textbf{Conversational Document Prediction} (\textit{CDP}) can be viewed as a conversational search problem, where the entire conversation context or a subset of it could be used as the query for retrieving matching documents. Compared to ad-hoc retrieval settings, using a conversational interface, the agent/system can ask clarification questions and interactively modify the search results as the conversation progresses \cite{zhang2018towards, aliannejadi2019asking}. 

The CDP task has been primarily addressed so far using ``traditional''  information retrieval (IR) techniques. 
\citet{habibi2015keyword} proposed a document recommender system by extracting keywords from a conversation using topic modeling techniques. 
\citet{ferreira2019classy} have used a similar keyword extraction framework and reported their results on a proprietary dataset. 

Many aspects of IR systems have undergone a revolution with the advent of powerful Deep Learning (DL) techniques in recent years \cite{mitra2018introduction, yang2019simple}. Yet this superior performance comes with high demand in computational resources as well as longer inference times, which hinders their application in real-world IR systems. Thus, the attention has been focused on techniques that reduce the computation complexity at the run-time without hindering the performance \cite{reimers2019sentence, lu2020twinbert}. 

In this work, we formulate the CDP task to support CC agents. We further 
release a new public dataset which enables research on the aforementioned task and investigate the performance of state-of-the art DL and IR models side-by-side on a number of datasets. We also analyze the runtime complexity of such systems, and propose a hybrid solution which is applicable in real-life systems.  

\section{Data}
\label{dataset}

We explore the CDP task using three datasets which contain human-to-human conversations between users and CC agents. Two of these datasets are internal: one from an internal customer support service on Mac devices (\textit{Mac-Support}) and another from an external client in the telecommunication domain (\textit{Telco-Support}). We also release a new \textit{Twitter} dataset, containing conversations between users and CC agents in 25 organizations on the Twitter platform\footnote{The Twitter dataset is available at:\\ \url{https://github.com/IBM/twitter-customer-care-document-prediction}}. We summarize the statistics of the three datasets in Table \ref{tab:data:overall}.

For our internal datasets, we filter out dialogs where: a) the agent doesn't provide a URL to the user, b) the URL is not in-domain (e.g.\,  Google searches, Microsoft forums, etc.), and focus on URLs from internal customer service knowledge base, and c) the URL is either no longer valid or has no content (e.g., login page). 
For Twitter dataset, we used the user\_timeline API to collect the tweets from agents containing in-domain URLs. The dialogs were constructed starting from these tweets and identifying the previous user and agent tweets to these tweets. If a dialog contains multiple URLs, we only use the dialog till the first agent utterance containing a URL. The details for document content extraction are in Appendix.

From Table \ref{tab:data:overall}, we observe that, around 5-9\% dialogs include a URL document provided by the agent. We also note that the website content for organizations gets updated frequently as many URLs return 404 errors. The average number of turns in a dialog and dialog length (in tokens) is much smaller for \textit{Twitter} in comparison to the \textit{Mac-Support} and \textit{Telco-Support} datasets. Our experiments results in Section \ref{results}, particularly BM25 and IRC in Table \ref{tab:results:all-datasets}, demonstrate the importance of dialog context for the CDP task, even when that context is not very rich, as is the case for the short dialogs of Twitter.

\section{Approaches}
\label{approaches}
We now formally introduce the CDP task and notations below. We then describe two alternative approaches (IR and DL) and their hybrid that we evaluate for this task.  

\subsection{Task Definition}
We regard the CDP task as a dialogue-based document classification task, similar to next utterance classification \cite{loweubuntu}. This is achieved by processing the data as described in Section \ref{neural-approaches}, without requiring any human labels. 

Formally, let $d=\{s_1 \colon t_1,s_2 \colon t_2,\ldots,s_n \colon t_n\}$ denote an $n$-turn dialog, where $s_i$ represents the speaker (\emph{user - U} or \emph{agent - A}), and $t_i$ represents the $i^{th}$ utterance. The dialog history is concatenated together to form dialog context of length $m$, represented as $d = (d_1, d_2, ..., d_i, ..., d_m)$, where $d_i$ is the $i^{th}$ word in context. Let $Y$ denote the set of all documents which can be recommended to the user. Similar to dialogs, each document $y\in{Y}$ is represented as 
$y = (y_1, y_2, ..., y_j, ..., y_n)$, where $y_j$ is the $j^{th}$ word in the document. 
Given dialog query $d$, the goal of the CDP task is to recommend $k$ documents in $Y$ to the agent. 
For evaluation, we use Recall@$k$ and Mean Reciprocal Rank, where the model is asked to select the $k$ most likely documents, and it is correct if the correct URL document is among these k documents.

\subsection{Information Retrieval approaches}
Following previous works, the first approach we evaluate for this task is based on IR models. We use an Apache 
Lucene index, employed with English language analyzer and default BM25 similarity~\cite{RobertsonBM25}. Documents in the index are represented using two fields. The first field contains the actual document content. The second field augments the document's representation with the text of all dialogs that link to it in the train-set~\cite{amitayQanchors}.

For a given (dialog) query $d$, matching documents are retrieved using four different ranking steps, which are combined using a cascade approach~\cite{Lidan2011}. Following~\cite{Gysel2016}, we obtain an initial pool of candidate documents using a lexical query aggregation approach. To this end, each utterance $t_i\in{d}$ is represented as a separate weighted query-clause, having its weight assigned relatively to its sequence position in the dialog~\cite{Gysel2016}. Various sub-queries are then combined using a single disjunctive query. The second ranker evaluates each document $y$ obtained by the first ranker against an expanded query (applying relevance model~\cite{Lavrenko2001}). The third ranker applies a manifold-ranking approach~\cite{Bin2011}, aiming to score content-similar documents (measured by Bhattacharyya language-model based similarity) with similar scores. 

The last ranker in the cascade treats the dialog query $d$ as a verbose query and applies the \textit{Fixed-Point} (FP) method~\cite{Paik2014} for weighting its words. Yet, compared to ``traditional'' verbose queries, dialogs are further segmented into distinct utterances. Using this observation, we implement an \textit{utterance-biased} extension for enhanced word-weighting. To this end, we first score the various utterances based on the initial FP weights of words they contain and their relative position. We then propagate utterance scores back to their associated words. The IR model is denoted as \emph{IRC}, short for \emph{IR-Cascade} in Table \ref{tab:results:all-datasets}.

\subsection{Neural approaches}
\label{neural-approaches}

The second type of approaches we evaluate are neural models. We process the datasets to construct triples of \textless dialog context ($d$), URL document content ($y$), label ($1/0$)\textgreater from each dialog.
For each $d$, we create a set of $k+1$ triples: one triple containing the correct URL provided by the agent (label - $1$), and $k$ triples containing incorrect URLs randomly sampled from $Y$ (label - $0$). We explore different values for $k$ and share additional results in Appendix.
During evaluation, we evaluate a given dialog context against the set of all documents ($Y$). 

We evaluate the CDP task using three state-of-the-art neural models: \textit{Enhanced Sequential Inference Model} (ESIM) proposed by \citet{chen2017enhanced} which performs well on Natural Language Inference (NLI) and next utterance prediction tasks \cite{dong2018enhance}, \textit{BertForSequenceClassification} model~\cite{Wolf2019HuggingFacesTS} and \textit{SBERT}. We next briefly describe the details for these models.

\subsubsection{ESIM}

The ESIM model takes two input sequences: dialog context ($d$) and document content ($y$), and feeds them through BiLSTM to generate local context-aware word representations denoted by $\bar{d}$ and $\bar{y}$. A co-attention matrix $E$, where $E_{ij} =\bar{d}_i^T\bar{y}_j$, computes the similarity between $d$ and $y$. The attended dialog context and document content vectors denoted by $\tilde{d}$ and $\tilde{y}$ are computed using $E$, which represent the most relevant word in $y$'s content for each word in $d$'s context and vice-versa.

This local inference information is enhanced by computing the difference and the element-wise product for the tuple \textless $\bar{d}, \tilde{d}$\textgreater\ as well as for \textless $\bar{y}, \tilde{y}$\textgreater. The difference and element-wise product are then concatenated with the original vectors, $\bar{d}$ and $\tilde{d}$ or $\bar{y}$ and $\tilde{y}$ respectively. The concatenated vectors are then fed to another set of BiLSTMs to compose the overall inference between the two sequences. Finally, the result vectors are converted to a fixed-length vector by max pooling and fed to a final classifier. 

\subsubsection{BERT}
We use pre-trained BERT~\cite{devlin2019bert} in two settings: a) fine-tuned on the training set, and b) an additional pre-training step on unlabeled data (dialogs in the training set and all documents) followed by fine-tuning on the training set (denoted as BERT$^*$ in Table \ref{tab:results:all-datasets}). In both settings, evaluation is done on the test set. 

We utilize the binary classifier (\textit{BertForSequenceClassification}) of BERT, commonly used for GLUE tasks 
\cite{wang2018glue} 
as follows.
A dialog context $d$ and a document $y$ are fed together to BERT as a sequence ([CLS] $d$ [SEP] $y$ [SEP]).
To adapt to BERT's limitation of maximum sequence length, we use 512 tokens and feed BERT with the 256 tokens each from $d$ and $y$, decided by a heuristic explained in Appendix. 
We use the hidden state of the [CLS] token as the representation of the pair. Training is done using positive and negative examples (Sec.~\ref{neural-approaches}) with cross-entropy loss. Re-ranking of candidate documents $\{y \in Y\}$ for a given context $d$ is done through the confidence score of each pair ($d$,$y$) which belongs to the positive class.  

\begin{table}
  \begin{center}
  \small
  \begin{tabular}{lcc}
    \toprule
    Model & \# param. & Inf. time (sec.) \\
    \cmidrule(r){2-3}
    BM25 & 2 & 0.02 \\
    IRC & 12 & 0.03 \\
    ESIM & 3.7M & 2.37 \\
    BERT($^*$) & 110M & 0.95 \\
    SBERT($^*$) & 110M & 0.04 \\
    \bottomrule
  \end{tabular}
  \end{center}
  \caption{Inference time for a single query from Twitter test set on a V100-PCIE-32GB GPU}
  \label{tab:results-inference-time}
\end{table}

\subsubsection{Sentence-BERT (SBERT)}
We also explore SBERT \cite{reimers2019sentence}, which uses a Siamese network structure to fine-tune the pre-trained BERT network and derive semantically meaningful sentence embeddings. The sentence embeddings for $d$ and $y$ are derived by adding a pooling operation (default: mean) on the BERT outputs and then can be compared using cosine-similarity to achieve low inference time. We fine-tune SBERT in the same two settings as BERT mentioned above. The input handling and evaluation is same as BERT above. 
The fine-tuning and hyperparameter details are available in Appendix.

\subsection{An hybrid approach}
To investigate the real-world use of our approaches, we compare (in Table \ref{tab:results-inference-time}) the number of parameters of each model and inference time for a single query from the Twitter test set. 
The IRC model is much faster in comparison to the neural models. For incorporating the additional performance gain from neural models (in Table \ref{tab:results:all-datasets}), we introduce an hybrid approach by a two-stage pipeline where we utilize the IRC model to generate a ranking of the document pool $Y$. The top-$k$ documents ($k: 20$) are then re-ranked through ESIM and recommended to the CC agent. This hybrid approach (IRC+ESIM) combines the best of both worlds. 

\begin{table}[t]
 \small
\begin{center}
\begin{tabular}{lcccc}
    \hline
    \textbf{Model} & \textbf{R@1} & \textbf{R@2} & \textbf{R@5} & \textbf{R@10}\\ \hline
    \multicolumn{5}{c}{\textit{Mac-Support}}\\\hline
    BM25 & 0.199 & 0.278 & 0.394 & 0.479\\ \hline
    IRC & 0.411 & 0.567 & 0.734 & 0.809 \\ \hline
    ESIM & 0.419 & 0.602 & 0.758 & 0.848 \\ \hline
    BERT & 0.319 & 0.441 & 0.655 & 0.809 \\ \hline
    BERT* & 0.315 & 0.447 & 0.698 & 0.818\\ \hline
    SBERT & 0.096 & 0.177 & 0.299 & 0.460\\ \hline
    SBERT* & 0.128 & 0.203 & 0.319 & 0.496\\ \hline\hline
    IRC+ESIM & \textbf{0.496} & \textbf{0.684} & \textbf{0.872} & \textbf{0.985} \\ \hline
    
    \multicolumn{5}{c}{\textit{Telco-Support}}\\\hline
    BM25 & 0.032 & 0.068  &  0.182 & 0.313 \\ \hline
    IRC & 0.405 & 0.551 & 0.735  & 0.867  \\ \hline
    ESIM & 0.676 & 0.806 & 0.911 & 0.951\\ \hline
    BERT & 0.523 & 0.699 & 0.866 & 0.918\\ \hline
    BERT* & 0.569 & 0.748 & 0.891 & 0.927\\ \hline
    SBERT & 0.250 & 0.391 & 0.612 & 0.758\\ \hline
    SBERT* & 0.360 & 0.506 & 0.711 & 0.826\\ \hline\hline
    IRC+ESIM & \textbf{0.721} & \textbf{0.863} & \textbf{0.942} & \textbf{0.964}\\ \hline
    
    \multicolumn{5}{c}{\textit{Twitter}}\\\hline
    BM25 & 0.088 & 0.150 & 0.224  & 0.306   \\ \hline
    IRC &  0.420 & 0.554  & 0.728  &  0.802 \\ \hline
    ESIM & 0.474  & 0.590  & 0.680 & 0.772 \\ \hline
    BERT & 0.400 & 0.418 & 0.424 & 0.428\\ \hline
    BERT* & 0.370 & 0.382 & 0.386 & 0.386\\ \hline
    SBERT & 0.182 & 0.246 & 0.354 & 0.442\\ \hline
    SBERT* & 0.224 & 0.308 & 0.484 & 0.644\\ \hline\hline
    IRC+ESIM & \textbf{0.559} & \textbf{0.684} & \textbf{0.819} & \textbf{0.902} \\ \hline

\end{tabular}
\end{center}
\caption{Performance of models on the test set of three datasets. R@$k$ refers to Recall at position $k$. MRR and corresponding validation results are in Appendix.} 
\label{tab:results:all-datasets}
\end{table}

\section{Results and Analysis}
\label{results}

Results are presented in Table \ref{tab:results:all-datasets}. We provide training setting and hyperparameter details for all neural models in Appendix. We observe that the ESIM model performs best across all datasets and the IRC model performs comparably to the ESIM model except for Telco-Support dataset. We observe a significant performance reduction with BERT models in comparison to both IRC and ESIM models. The BERT$^*$ model (additional pre-training) improves performance for Telco-Support dataset, but is still inferior to ESIM model. The SBERT models provide the benefit of low inference time, but reduce performance further. We conclude that for CDP task, explicit cross-attention between dialog context $d$ and document $y$ present in ESIM is crucial. The BERT models try to incorporate cross-attention through self-attention on the concatenated \textless $d, y$\textgreater\ pair sequence, but still lag behind.

Finally, the hybrid approach (IRC+ESIM) provides a significant boost in performance (e.g., between +7\%-20\% in R@1), and reduces the inference time of ESIM. This demonstrates the benefit and importance of combining IR models that are based on exact matching, with neural models that further allow semantic inference in the domain for real-world applications.

\section{Conclusion and Future Work}
We introduced the Conversational Document Prediction (CDP) task and investigated the performance of state-of-the-art DL and IR models. We also release a new public Twitter dataset on the CDP task. In this work, we considered only URL documents with content. Other potential document types that could be considered are PDFs, doc etc. and URLs without content (e.g. login, tracking). We plan to address these challenges in future work.

\balance

\bibliography{emnlp2020}
\bibliographystyle{acl_natbib}

\appendix

\vspace{0.6cm}
\section{Appendix: Additional results}
\label{full-validation-and-test-results}
The results for corresponding validation performance for  Information Retrieval and Neural approaches, as well as Mean Reciprocal Rank (MRR) metric for both validation and test sets on all datasets are available in Table \ref{tab:results:all-datasets-valid-and-test}.

\subsection{Negative samples for neural approaches}
\label{appendix-num-negative-samples-results}
For creating training data for our neural approaches, we create $k$ triples containing incorrect URLs sampled randomly from set of all documents. We experimented with different values for the hyper-parameter \emph{num\_negative\_samples} used for generating the training data. The results for ESIM model for the Mac-Support dataset are presented in Table \ref{tab:results:num-negative-samples-esim-mac-ibm}. We observe that increasing the number of negative samples doesn't improve the ESIM model performance significantly and \emph{num\_negative\_samples - 4} provides us the best of both worlds, i.e. good performance and lower training time, in comparison to using a higher negative sample ratio. We use the same value for all neural models for all datasets.

\subsection{Input handling for BERT models}
\label{appendix-bert-input-handling}
To handle the BERT model input limitation of 512 tokens max sequence length, we feed BERT with 256 tokens each from dialog context $d$ and document content $y$. We observe that the initial sentences in a URL document always capture the core gist of the document, so we always use the first 256 tokens from the document content. For dialog context, we observe that as the conversation progresses over multiple turns and the user query gets more complex, the conversation shifts from the original query to another problem in many dialogs. We explore two input approaches for deciding which tokens to consider if dialog context sequence length $|d| > 256$:
\begin{enumerate}
    \item Input-A: Truncate the dialog context $d$ to consider only the first 256 tokens from the dialog context.
    \item Input-B: Ignore tokens in the middle of dialog context sequence to reduce the $|d|$ to 256.
\end{enumerate}
The results for both approaches for BERT model on the Telco-Support dataset are in Table \ref{tab:results:input-handling-bert-telco}. We use the same heuristic for all neural models for all datasets.

\begin{table}[ht]
\begin{center}
\begin{tabular}{lccccc}
    \hline
    \multicolumn{1}{p{1cm}}{\centering negative\\samples} & 
    \multirow{2}{*}{R@1} &
    \multirow{2}{*}{R@2} &
    \multirow{2}{*}{R@5} &
    \multirow{2}{*}{R@10} &
    \multirow{2}{*}{MRR} \\ \hline

    4 & 0.417 & 0.535 & 0.676 & 0.745 & 0.534 \\ \hline
    7 & 0.400 & 0.507 & 0.633 & 0.728 & 0.510 \\ \hline
    10 & 0.419 & 0.509 & 0.683 & 0.779 & 0.534 \\ \hline
    14 & 0.419 & 0.518 & 0.678 & 0.747 & 0.531 \\ \hline
    
\end{tabular}
\end{center}
\caption{Performance of ESIM model on the validation set of Mac-Support dataset for different values of num\_negative\_samples.}
\label{tab:results:num-negative-samples-esim-mac-ibm}
\end{table}

\begin{table}[ht]
\begin{center}
\begin{tabular}{lccccc}
    \hline
    \multicolumn{1}{p{1cm}}{\centering dialog\\input} & 
    \multirow{2}{*}{R@1} &
    \multirow{2}{*}{R@2} &
    \multirow{2}{*}{R@5} &
    \multirow{2}{*}{R@10} &
    \multirow{2}{*}{MRR} \\ \hline

    Input-A & 41.19 & 58.72 & 80.03 & 88.8 & 57.26 \\ \hline
    Input-B & 52.38 & 70.89 & 87.13 & 92.4 & 67.04 \\ \hline

\end{tabular}
\end{center}
\caption{Performance of BERT model on the validation set of Telco-Support dataset for dialog context sequence input handling.}
\label{tab:results:input-handling-bert-telco}
\end{table}

\begin{table*}[ht]
\begin{center}
\begin{tabular}{lccccc|ccccc}
    \hline
    \multirow{2}{*}{\textbf{Model}} & \multicolumn{5}{c}{\textbf{Validation}} & \multicolumn{5}{c}{\textbf{Test}}\\
     & R@1 & R@2 & R@5 & R@10 & MRR & R@1 & R@2 & R@5 & R@10 & MRR  \\ \hline
    \multicolumn{11}{c}{\textit{Mac-Support}}\\\hline
    BM25 & 0.169 & 0.236 &  0.358 & 0.437 & 0.246  & 0.199 & 0.278 & 0.394 & 0.479 & 0.283  \\ \hline
    IRC & 0.407 & 0.548 & 0.713 & 0.805 &  0.537 & 0.411 & 0.567 & 0.734 & 0.809 & 0.546 \\ \hline
    ESIM & 0.417 & 0.569 & 0.741 & 0.831 & 0.561 & 0.419 & 0.602 & 0.758 & 0.848 & 0.573 \\ \hline
    BERT & 0.302 & 0.447 & 0.625 & 0.769 & 0.444 & 0.319 & 0.441 & 0.655 & 0.809 & 0.470 \\ \hline
    BERT* & 0.332 & 0.501 & 0.711 & 0.824 & 0.497 & 0.315 & 0.447 & 0.698 & 0.818 & 0.471 \\ \hline
    SBERT  & 0.079 & 0.137 & 0.267 & 0.434 & 0.185 & 0.096 & 0.177 & 0.299 & 0.460 & 0.207 \\ \hline
    SBERT* & 0.100 & 0.149 & 0.312 & 0.494 & 0.214 & 0.128 & 0.203 & 0.319 & 0.496 & 0.238 \\ \hline

    \multicolumn{11}{c}{\textit{Telco-Support}}\\\hline
    BM25 & 0.039  & 0.069  & 0.178 & 0.313 & 0.100 & 0.032 & 0.068  &  0.182 & 0.313  & 0.097 \\ \hline
    IRC & 0.409  &  0.549 & 0.737  & 0.859 & 0.547 & 0.405 & 0.551 & 0.735  & 0.867  & 0.546 \\ \hline
    ESIM & 0.683 & 0.803 & 0.913 & 0.953 & 0.782 & 0.676 & 0.806 & 0.911 & 0.951 & 0.779 \\ \hline
    BERT & 0.524 & 0.709 & 0.871 & 0.924 & 0.670 & 0.523 & 0.699 & 0.866 & 0.918 & 0.667\\ \hline
    BERT* & 0.568 & 0.745 & 0.899 & 0.936 & 0.706 & 0.569 & 0.748 & 0.891 & 0.927 & 0.702\\ \hline
    SBERT & 0.266 & 0.403 & 0.626 & 0.760 & 0.423 & 0.250 & 0.391 & 0.612 & 0.758 & 0.410 \\ \hline
    SBERT* & 0.365 & 0.518 & 0.724 & 0.834 & 0.521 & 0.360 & 0.506 & 0.711 & 0.826 & 0.512   \\ \hline
    
    \multicolumn{11}{c}{\textit{Twitter}}\\\hline
    BM25 & 0.111  & 0.156  & 0.265  & 0.375 & 0.177 & 0.088 & 0.150 & 0.224  & 0.306  & 0.148 \\ \hline
    IRC & 0.499 & 0.625  & 0.777  &  0.819 &  0.611 & 0.420 & 0.554  & 0.728  &  0.802 &  0.549  \\ \hline
    ESIM & 0.548 & 0.642 & 0.747 & 0.806 & 0.642 & 0.474  & 0.590  & 0.680 & 0.772 & 0.579 \\ \hline
    BERT & 0.474 & 0.489 & 0.489 & 0.489 & 0.482 & 0.400 & 0.418 & 0.424 & 0.428 & 0.411 \\ \hline
    BERT* & 0.474 & 0.484 & 0.484 & 0.484 & 0.479 & 0.370 & 0.382 & 0.386 & 0.386 & 0.377\\ \hline
    SBERT & 0.226 & 0.295 & 0.373 & 0.447 & 0.311 & 0.182 & 0.246 & 0.354 & 0.442 & 0.276  \\ \hline
    SBERT* & 0.321 & 0.417 & 0.573 & 0.683 & 0.439 & 0.224 & 0.308 & 0.484 & 0.644 & 0.349  \\ \hline
    
\end{tabular}
\end{center}
\caption{Performance of models on the validation and test sets for the three datasets. R@$k$ refers to Recall at position $k$ in all documents, denoted as R@1, R@2, R@5 and R@10. MRR refers to the Mean Reciprocal Rank.}
\label{tab:results:all-datasets-valid-and-test}
\end{table*}

\section{Appendix: Model Training and Hyperparameter Details}
\subsection{ESIM model}
\label{esim-training-details-appendix}
We used 300-dimensional Glove pre-trained vectors \cite{pennington2014glove}, 100-dimensional word2vec vectors \cite{mikolov2013efficient} and 80-dimensional character embedding vectors for generating the word representation. For training word2vec vectors, we use the \texttt{gensim} API with the following hyper-parameters: size=100, window=10, min\_count=1 and epochs=20. We also incorporate character embeddings to our ESIM implementation \cite{dong2018enhance}. The final prediction layer is a 2-layer fully-connected feed-forward neural network with ReLu activation. We use sigmoid function and minimize binary cross-entropy loss for training and updating the model. We used Adam \cite{kingma2014adam} with a learning rate of 0.001 and exponential decay with a decay rate of 0.96 decayed every 5000 steps. The number of hidden units for BiLSTMs was 256. For the prediction layers, we used 256 hidden units with ReLU activation.

\subsection{Additional pretraining for BERT model}
\label{bert-lm-pretraining-details-appendix}
We use the BERT-Base, Uncased model from \href{https://storage.googleapis.com/bert_models/2018_10_18/uncased_L-12_H-768_A-12.zip}{BERT-Base-Uncased} - 12-layer, 768-hidden, 12-heads, 110M parameters - as the base model for our experiments. For convenience, we refer to BERT-Base-Uncased as \$BERT below. We use the code from \href{https://github.com/google-research/bert}{Google-Research Bert Github} for creating pretraining data as well as to run additional pretraining on our domain data. We only use the training dialogs and contents from all documents for creating pretraining data. The hyperparameters used for creating pretraining data are:

\begin{verbatim}
vocab_file=$BERT/vocab.txt
do_lower_case=True
max_seq_length=512
max_predictions_per_seq=20
masked_lm_prob=0.15
random_seed=12345
dupe_factor=10
\end{verbatim}

The hyperparameters used to run LM-pretraining are:
\begin{verbatim}
train_batch_size=16
max_seq_length=512
max_predictions_per_seq=20
num_train_steps=100000
num_warmup_steps=10000
save_checkpoints_steps=20000
learning_rate=5e-5
\end{verbatim}

\subsection{Fine-tuning BERT model}
\label{bert-fine-tuning-details-appendix}
The hyperparameters used for further fine-tuning BERT model are:
\begin{verbatim}
do_lower_case=True
max_seq_length=512
per_gpu_eval_batch_size=24
per_gpu_train_batch_size=24
learning_rate=2e-5
num_train_epochs=5
\end{verbatim}
The model is periodically evaluated on the validation set after $n$ steps, which is decided based on the training dataset size.

\subsection{Fine-tuning SBERT model}
The hyperparameters used for fine-tuning SBERT model are:
\begin{verbatim}
do_lower_case=True
max_seq_length=256
batch_size=16
learning_rate=2e-5
num_train_epochs=5
optimizer=Adam 
\end{verbatim}
We use a linear learning rate warm-up over 10\% of the training data. We fine-tune SBERT with a 3-way softmax-classifier  objective function and the default pooling strategy is MEAN. The max\_seq\_length is 256 each for dialog context $d$ and document content $y$. For SBERT*, we use the same additional pre-trained BERT* model from before. The model is periodically evaluated on the validation set after $n$ steps, which is decided based on the training dataset size.

\section{Appendix: Extracting content from URL documents}
For the internal Mac-Support dataset, the document content for each URL was obtained by API calls to the customer service knowledge base. For the Telco-Support and Twitter datasets, we capture the HTML content using a Selenium Chrome webdriver, which renders the URL document by loading all CSS styling and Javascript. The extracted HTML was cleaned through a Markdown generation pipeline, where we manually identify and filter the DOM tags (using CSS id and/or class) which correspond to header(s), footer, navigation bars etc. This process is repeated for each URL domain in both datasets. The tools for data preprocessing are available here: \url{https://github.com/IBM/MDfromHTML}.

\end{document}